\documentclass[sigconf]{acmart}
\AtBeginDocument{%
  \providecommand\BibTeX{{%
    \normalfont B\kern-0.5em{\scshape i\kern-0.25em b}\kern-0.8em\TeX}}}

\copyrightyear{2021} 
\acmYear{2021} 
\setcopyright{acmcopyright}\acmConference[MM '21]{Proceedings of the 29th ACM International Conference on Multimedia}{October 20--24, 2021}{Virtual Event, China}
\acmBooktitle{Proceedings of the 29th ACM International Conference on Multimedia (MM '21), October 20--24, 2021, Virtual Event, China}
\acmPrice{15.00}
\acmDOI{10.1145/3474085.3475280}
\acmISBN{978-1-4503-8651-7/21/10}

\usepackage{subfigure}
\usepackage{algorithmicx,algorithm}
\usepackage[noend]{algpseudocode}
\usepackage{color}
\usepackage{url}
\usepackage{balance}

\acmSubmissionID{609}

\begin{document}
\fancyhead{}
\title{Imitating Arbitrary Talking Style for Realistic Audio-Driven Talking Face Synthesis}

\author{Haozhe Wu$^{1}$,\quad Jia Jia$^{1}$*,\quad Haoyu Wang$^{1}$,\quad Yishun Dou$^{2}$,\quad Chao Duan$^{2}$,\quad Qingshan Deng$^{2}$}

\makeatletter
\def\authornotetext#1{
\if@ACM@anonymous\else
    \g@addto@macro\@authornotes{
    \stepcounter{footnote}\footnotetext{#1}}
\fi}
\makeatother
\authornotetext{Corresponding author.}

\affiliation{
 \institution{\textsuperscript{\rm 1}Department of Computer Science and Technology, Tsinghua University, Beijing 100084, China}
 \institution{Beijing National Research Center for Information Science and Technology (BNRist)}
 \institution{The Institute for Artificial Intelligence, Tsinghua University}
 \institution{\textsuperscript{\rm 2}HiSilicon Company, Shenzhen, China}
 wuhz19@mails.tsinghua.edu.cn,  jjia@tsinghua.edu.cn, wang-hy18@mails.tsinghua.edu.cn,
\{douyishun,duanchao15,dengqingshan\}@hisilicon.com
 }

\def\authors{Haozhe Wu, Jia Jia, Haoyu Wang, Yishun Dou, Chao Duan, Qingshan Deng}

\renewcommand{\shortauthors}{Wu et al.}


\begin{abstract}
People talk with diversified styles. %
For one piece of speech, different talking styles exhibit significant differences in the facial and head pose movements. %
For example, the "excited" style usually talks with the mouth wide open, while the "solemn" style is more standardized and seldomly exhibits exaggerated motions. %
Due to such huge differences between different styles, %
it is necessary to incorporate the talking style into audio-driven talking face synthesis framework. %
In this paper, we propose to inject style into the talking face synthesis framework through imitating arbitrary talking style of the particular reference video. %
Specifically, we systematically investigate talking styles with our collected \textit{Ted-HD} dataset and construct style codes as several statistics of 3D morphable model~(3DMM) parameters. Afterwards, we devise a latent-style-fusion~(LSF) model to synthesize stylized talking faces by imitating talking styles from the style codes. %
We emphasize the following novel characteristics of our framework: (1) It doesn't require any annotation of the style, the talking style is learned in an unsupervised manner from talking videos in the wild. %
(2) It can imitate arbitrary styles from arbitrary videos, and the style codes can also be interpolated to generate new styles. %
Extensive experiments demonstrate that the proposed framework has the ability to synthesize more natural and expressive talking styles compared with baseline methods. %

\end{abstract}

\ccsdesc[300]{Computing methodologies~Animation}

\keywords{photorealistic talking face, talking style, style imitation}


\maketitle

\section{Introduction}

\begin{figure}[h]
  \centering
  \includegraphics[width=0.9\linewidth]{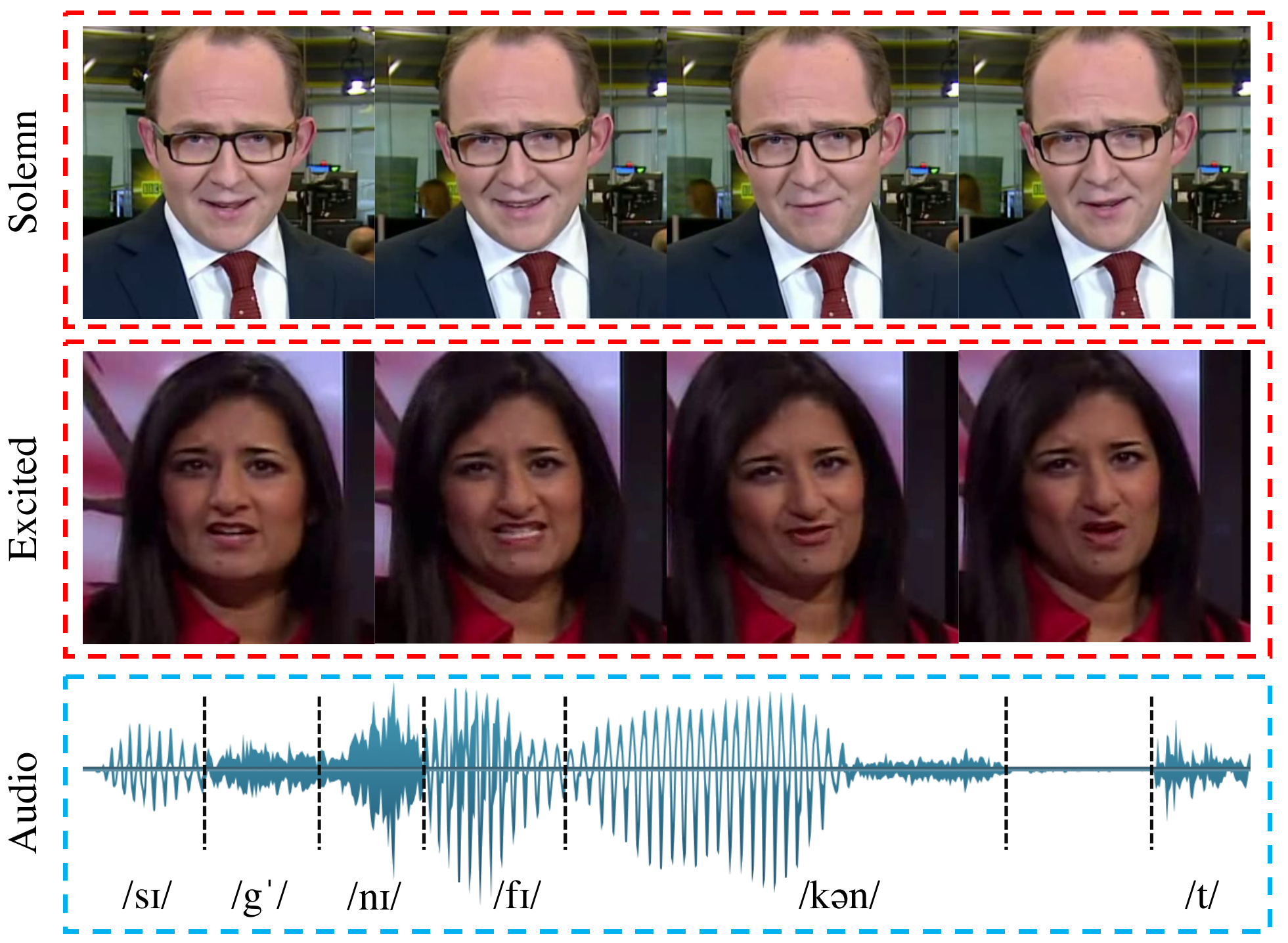}
  \caption{Different talking styles have significant differences in facial and head pose movements when pronouncing the word "significant". }
  \label{fig:intro}
\end{figure}

The talking face synthesis is an eagerly anticipated technique for several applications, %
such as filmmaking, teleconference, virtual/mix reality, and human-computer interaction. %
One of the key essences behind the talking face synthesis is the stylization of facial and head pose movements. %
Different from the talking emotion which reflects in the short-term facial motions, the talking style is a crucial factor which affects long-term facial and head pose movements. %
People usually talk with diversified talking styles such as "excited", "solemn", "communicational", "storytelling", \textit{et al}. %
Given one piece of speech, different talking styles exhibit significant differences in the facial and head pose movements. %
For example, as shown in Figure~\ref{fig:intro}, people with the "excited" style usually talk aloud, and thus the facial movement of the mouth wide open occurs frequently. Meanwhile, the "solemn" talking style usually appears in formal occasions, and thus the exaggerated motion seldom occurs. %
Considering such huge differences between different styles, %
in order to synthesize diversified and realistic talking faces with respect to one piece of speech, %
it is necessary to incorporate talking style into the audio-driven talking face synthesis framework. %


\begin{figure*}[h]
  \centering
  \includegraphics[width=0.9\linewidth]{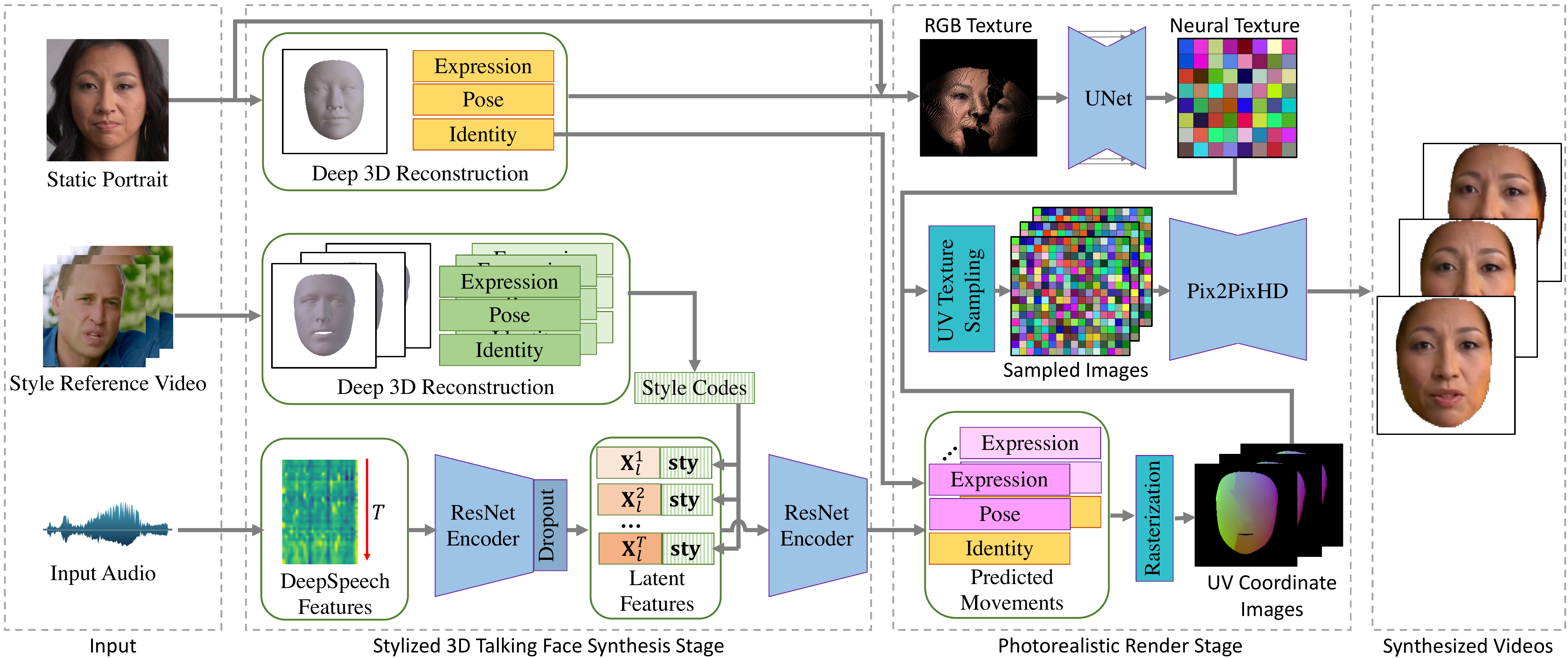}
  \caption{The overall framework of our method contains two stages: the stylized 3D talking face synthesis stage and the photorealistic render stage. The first stage synthesizes stylized 3D talking faces by imitating talking styles from the style codes. The second stage synthesizes photorealistic videos with deferred neural render and neural texture generation model. The stylized 3D talking face synthesis stage and the photorealistic render stage are trained separately in our framework. $T$ denotes the time dimension of DeepSpeech features. For the calculation of style codes, \textit{even  style reference videos from one identity would yield diversified style codes}. }
  \label{fig:framework}
\end{figure*}

Previous efforts~\cite{cudeiro2019capture,yi2020audio} have shown the rationality of synthesizing stylized talking faces. %
Yi~\textit{et al.}~\cite{yi2020audio} proposed Memory Augmented GAN model to synthesize stylized talking face with the wild training data. %
Cudeiro~\textit{et al.}~\cite{cudeiro2019capture} proposed Voice Operated Character Animation~(VOCA) model to learn the identity-wise talking style. %
The VOCA model captures the talking style of each identity by injecting the one-hot identity vector into the audio-motion predicting network. %
Overall, these methods have the following two disadvantages which constrain the expressiveness of the synthesized talking styles: %
(1) These methods assume that each identity has only one talking style. However, in the real scenario, the talking style is only relatively stable inside one video clip, one person can talk with significantly different styles among different video clips. %
(2) These methods require substantial labour to collect adequate synchronized audio-visual data for each identity, which is unavailable for the wild scenarios. %

To address the aforementioned problems, we propose to synthesize stylized talking faces by imitating styles from  arbitrary clips of videos. %
With such motivation, the talking videos with stable and diversified talking styles are in need. %
Therefore, we collect the \textit{Ted-HD} dataset which has 834 clips of videos with 60 identities. %
Each video clip has an average length of 23.5 seconds. %
The talking styles of the \textit{Ted-HD} dataset are diversified, and the talking style inside each video clip is stable. %

Based on the constructed dataset, we devise a two stage talking face synthesis framework as shown in Figure~\ref{fig:framework}. 
The first stage imitates talking styles from arbitrary videos and synthesizes 3D talking faces according to the driven speech. %
Afterwards, in the second stage, we render the 3D face model photo-realistically from one static portrait of the speaker. 
Overall, the key idea of our framework is to construct style codes from video clips in the wild, and then synthesize talking faces by imitating the talking styles from the constructed style codes. %
Specifically, for the style codes construction, %
we conduct exhaustive observations on the \textit{Ted-HD} dataset, which on one hand verify that even one identity has multiple talking styles, on the other hand find that the talking style is closely related to the variance of facial and head pose movements inside each video. %
With our observations, we define the style codes as several interpretable statistics of 3D morphable model~(3DMM)~\cite{blanz1999morphable} parameters. %
Having obtained the style codes of each talking video,  we devise a latent-style-fusion~(LSF) model to synthesize stylized 3D talking faces by imitating talking styles from the style codes. %
Detailedly, the LSF model first dropouts~\cite{srivastava2014dropout} information from the audio stream to prevent the audio from dominating the synthesis process. %
Further, the LSF model fuses the style codes with the latent audio representation 
frame-by-frame to synthesize 3D talking face with corresponding talking style. %
The overall implementation of the LSF model is simple but effective. %
Our model not only circumvents the annotation for talking styles and avoids collecting substantial training data for each identity, but also enables new talking style generation. %

With our proposed framework, talking faces with diversified styles can thus be synthesized. %
We perform experiments on the \textit{Ted-HD} dataset for evaluation. Compared with baseline methods, our framework synthesizes more expressive and diversified talking styles. %
We conduct extensive user studies to investigate the facial motion naturalness and audio-visual synchronization. %
With the mean opinion score~(MOS) of 20 participants, our framework outperforms baseline methods by 0.67 on average in terms of facial motion naturalness and 0.11 on average in terms of audio-visual synchronization. %

To summarize, our contributions are summarized as three-fold:

\begin{itemize}
    \item We propose to synthesize stylized talking faces by imitating talking styles of arbitrary videos. The incorporation of style imitation leads to more diversified talking styles. %
    \item We formalize the style codes of each talking video and devise the latent-style-fusion~(LSF) model to synthesize stylized 3D talking faces from style codes and driven audio. Our framework do not require any annotation of talking styles for stylized talking face synthesis. %
    \item We collect \textit{Ted-HD} dataset, which contains 834 clips of wild talking videos with stable and diversified talking styles. Based on the \textit{Ted-HD} dataset, we conduct extensive style observations and synthesize expressive talking styles. Code and dataset are publicly available~\footnote{\url{https://github.com/wuhaozhe/style_avatar}}. %
\end{itemize}

\section{Related Work}


Talking face synthesis has received significant attention in previous literatures. %
Related work in this area can be grouped into two categories: %
the unimodal talking face synthesis~\cite{thies2020neural,zhou2019talking,chen2018lip,wiles2018x2face,prajwal2020lip,jamaludin2019you,zhou2021pose} and the multimodal talking face synthesis~\cite{wang2020mead,emre2020speech,cudeiro2019capture,yi2020audio,zeng2020talking,ji2021audio}. %
For one piece of driven speech, the unimodal talking face synthesis generates unique motion, while the multimodal talking face synthesis generates diversified facial and head pose movements. %

Most of the prior works focused on the unimodal talking face synthesis. %
Karras~\textit{et al.}~\cite{karras2017audio} proposed to synthesize 3D talking face with driven audio and emotion state. %
Suwajanakorn~\textit{et al.}~\cite{suwajanakorn2017synthesizing} synthesized high quality videos of talking Obama through hours of Obama's weekly address footage. %
Since that Suwajanakorn's method requires hours of data for each identity, several methods~\cite{thies2020neural,zhou2020makelttalk,zhou2019talking,prajwal2020lip,chen2019hierarchical,chen2018lip,yu2020multimodal,chen2020talking} proposed to simultaneously reduce the required duration of training data and guarantee the photorealism of the synthesized video. %
The ATVG framework proposed by Chen~\textit{et al.}~\cite{chen2019hierarchical} and the DAVS framework proposed by Zhou~\textit{et al.}~\cite{zhou2019talking} synthesize talking face with only one image. %
Despite that these unimodal talking face synthesis methods can synthesize photo-realistic videos, the lack of style results in the stiffness of the synthesized results. %


To synthesize diversified facial and head pose movements, some recent literatures addressed the multimodal talking face synthesis. %
Wang~\textit{et al.}~\cite{wang2020mead} and Eskimez~\textit{et al.}~\cite{emre2020speech} achieved multimodal synthesis through incorporating emotion condition vector, which enables the generation of diversified facial expressions. %
However, the synthesized results of these methods still lack in personality because of the overlook of talking style. %
To address this issue, some methods~\cite{yi2020audio,cudeiro2019capture} proposed to incorporate talking style into the synthesis framework. %
Yi~\textit{et al.}~\cite{yi2020audio} proposed Memory-Augmented GAN model to synthesize stylized talking face with wild training data. %
However, only the head pose is synthesized with multiple styles, while the facial movements still lack personality. %
Further, Cudeiro~\textit{et al.}~\cite{cudeiro2019capture} proposed the Voice Operated Character Animation~(VOCA) model to learn the identity-wise talking style. %
The VOCA model injects one-hot identity vector into the audio-motion predicting network, leading to discriminative styles of facial and head pose movements. %
However, the VOCA model on one hand requires substantial data for each identity, on the other hand forces one identity to have only one talking style, limiting its capacity on synthesizing diversified styles. %
To resolve this problem, in this work, we propose to imitate talking styles from arbitrary wild talking videos. %


\section{Problem Formulation}

In this paper, we propose a two-stage talking face synthesis framework which synthesizes stylized talking videos with the following three inputs: one static portrait of the speaker, the driven audio and the style reference video. %
We formalize the first stage of the framework as the 3D talking face synthesis stage and the second stage as the photorealistic render stage. %
Between two stages, we apply the 3DMM face model~\cite{blanz1999morphable} as a crucial bridge. Therefore, before formally defining the two stages, we firstly give a brief introduction of the face model we use. %

We leverage the 3DMM face model to represent each 3D face. 
With 3DMM, the face shape $\mathbf{S}$ is represented as an affine model of facial expression and facial identity: %

\begin{equation}
  \mathbf{S} = \mathbf{S(\alpha, \beta)} = \mathbf{\bar{S}} + \mathbf{B}_{id}\mathbf{\alpha} + \mathbf{B}_{exp}\mathbf{\beta},
\end{equation}
where $\mathbf{\bar{S}} \in \mathbf{R}^{N \times 3}$ is the average face shape; N is the number of the vertexes in the face model; $\mathbf{B}_{id}$ and $\mathbf{B}_{exp}$ are the PCA bases of identity and expression; $\alpha$ and $\beta$ are the identity parameters and the expression parameters. %
Following deng~\textit{et al.}~\cite{deng2019accurate}, we adopt the 2009 Basel Face Model~\cite{paysan20093d} for $\mathbf{\bar{S}}$, $\mathbf{B}_{id}$, and use expression bases $\mathbf{B}_{exp}$ of Guo~\textit{et al.}~\cite{guo2018cnn} built from Facewarehouse~\cite{cao2013facewarehouse}, resulting in $\alpha \in \mathbb{R}^{80}$, $\beta \in \mathbb{R}^{64}$. %
Afterwards, the 3D face shape is projected on the 2D plane according to the head pose and translation $p \in \mathbb{R}^{7}$, where 4 elements represent the pose quaternion and 3 elements represent the translation. %
Overall, the parameter set $(\alpha, \beta, p)$ controls the appearance of each face. %
In our framework, the facial movements are the time series of parameter $\beta$, which we denote as $\beta(t)$, while the head pose movements are the time series of parameter $p$, which we denote as $p(t)$. %
With $\beta(t)$ and $p(t)$, we formalize the two stages of our framework as follows. %

\begin{figure*}[h]
  \centering
  \includegraphics[width=0.9\linewidth]{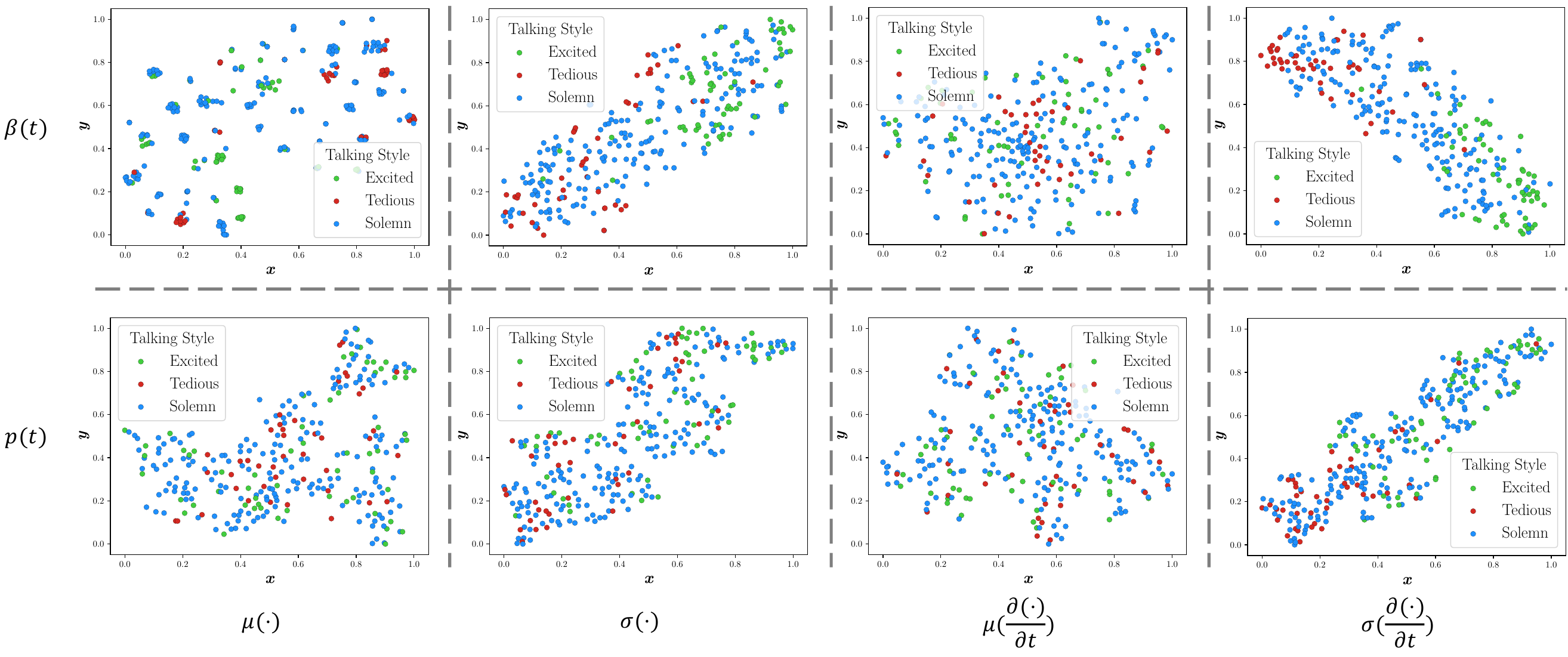}
  \caption{Correlation between talking styles and 3DMM parameter series. We investigate 8 different statistics and find that $\sigma(\beta(t)), \sigma(\frac{\partial \beta(t) }{\partial t }), \sigma(\frac{\partial p(t) }{\partial t })$ are mostly correlated with talking styles. }
  \label{fig:observe}
\end{figure*}

\textbf{3D Talking Face Synthesis Stage}. In this stage, given driven audio $\mathbf{X}_{a}$, the facial and head pose movements of style reference video $\beta_{sty}(t), p_{sty}(t)$, we aim to generate corresponding facial and head pose movements $\beta_{pred}(t)$, $p_{pred}(t)$. %

\textbf{Photorealistic Render Stage}. In this stage, given the predicted movements $\beta_{pred}(t)$, $p_{pred}(t)$ and the input portrait $\mathbf{X}_{p}$, we aim to generate photorealistic videos $\mathbf{Y}$. 



\section{Talking Style Observation}
\label{sec:obv}
In this section, we systematically investigate how different talking styles reflect in the facial and head pose movements $\beta(t), p(t)$. %
Afterwards, we formally define interpretable style codes for each video based on our observation. %

In order to observe talking styles of each video, %
we should firstly collect a suitable dataset for observation. %
The video for the style observation requires the following characteristics: %
(1) It should be of high resolution, (2) it should contain natural and expressive facial and head pose movements, %
(3) each clip of video cannot be too short, or the talking style can hardly be observed, %
(4) the talking style should both be stable inside the clip and be diversified across different clips, %
(5) the camera pose and location should be static inside each clip, otherwise the head pose parameters will be influenced by the camera movements. %
Considering the characteristics above, current publicly available wild datasets like VoxCeleb2~\cite{Chung18b} and LRS3~\cite{Afouras18d} are much too noisy and thus do not meet the requirements. %
Meanwhile, several in-lab datasets like MEAD~\cite{kaisiyuan2020mead} and GRID~\cite{cooke2006audio} do not have natural facial expressions and head poses, which dissatisfy the requirements either. %

To address this issue, we manually collect a wild dataset \textit{Ted-HD} which is suitable for style observation and further style synthesis. %
The \textit{Ted-HD} dataset selects several speech videos from Ted website. %
Each video in the dataset has one person giving a speech, %
which focuses on the facial part of each person and is of high resolution. %
We cut each video into several clips according to the scene change. %
Totally, \textit{Ted-HD} dataset contains 834 clips of videos with 60 identities. The average length of each clip is 23.5 seconds, and the total duration of the dataset is 6 hours. %
The talking style of these videos are diversified across different clips. %
Even for the same identity, there could be different talking styles. %

Having obtained the dataset, for each video, we reconstruct the facial and head pose movements $\beta(t), p(t)$. %
 We conduct exhaustive data observations on the correlation between talking style and $\beta(t), p(t)$, with the aim of answering the following questions: 

\begin{itemize}
    \item \textbf{Q1}: Whether one identity has multiple talking styles. %
    \item \textbf{Q2}: How the talking style is reflected in time series $\beta(t), p(t)$. %
\end{itemize}


To answer the \textbf{Q1}, we verify the talking style diversity of each identity through user study with A/B test. %
Specifically: we first randomly construct 100 triplets, each triplet contains two talking videos $v_{1}, v_{2}$ from the same identity and one talking video $v_{3}$ from the other identity. %
Next, we reconstruct $\beta(t), p(t)$ of $v_{1}, v_{2}, v_{3}$, retarget $\beta(t), p(t)$ to the same identity and render the retargeted faces to videos. The retargeted videos are signified as $v_{1}^{'}, v_{2}^{'}, v_{3}^{'}$. %
Afterwards, we show $v_{1}^{'}, v_{2}^{'}, v_{3}^{'}$ and their transcripts to users,
asking the following question: which pair of $(v_{1}^{'}, v_{2}^{'})$ and $(v_{1}^{'}, v_{3}^{'})$ has more similar talking styles. Statistics demonstrate that among 100 triplets, $(v_{1}^{'}, v_{3}^{'})$ is more similar in 30 triplets, while $(v_{1}^{'}, v_{2}^{'})$ is more similar in 70 triplets. %
With the statistics that $30\%$ of videos inside each identity have dissimilar talking styles, we conclude that one identity has multiple talking styles.


Since that the talking style is not consistent for each identity, formulating style codes is necessary. %
Therefore, in \textbf{Q2}, we conduct experiments to find how talking style is reflected in time series $\beta(t), p(t)$. %
Specifically, we first randomly select 300 talking videos and annotate the talking style of these videos to three categories: tedious, solemn, and excited. %
Afterwards, for the motion series $\beta(t), p(t)$ of each video, we calculate its derivative series with respect to time $t$: %
$(\frac{\partial \beta(t) }{\partial t }, \frac{\partial p(t) }{\partial t })$. %
Next, we calculate the mean value $\mu(\cdot)$ and the standard deviation $\sigma(\cdot)$ of $(\beta(t), p(t), \frac{\partial \beta(t) }{\partial t }, \frac{\partial p(t) }{\partial t })$ along time, yielding 8 feature vectors~(4 for mean, 4 for standard deviation). %
To observe the relationship between these feature vectors and talking styles, we leverage t-SNE algorithm~\cite{van2008visualizing} to visualize each feature vector and plot points from different style categories with different colors, as shown in Figure~\ref{fig:observe}. %
Figure~\ref{fig:observe} demonstrates that the talking style is closely related with $\sigma(\beta(t))$, $\sigma(\frac{\partial \beta(t) }{\partial t })$, $\sigma(\frac{\partial p(t) }{\partial t })$, especially $\sigma(\frac{\partial p(t) }{\partial t })$. %
Meanwhile the talking style is less related with $\mu(\cdot)$, %
which denotes that the talking style is mostly reflected in the fluctuation of movements, rather than the idle state of movements. %

Based on such observation, we define the style codes as the standard deviation of facial and head pose movements. %
Formally, given arbitrary video with the reconstructed parameter series $\beta(t), p(t)$, %
the style codes $\mathbf{sty}$ are defined as:
\begin{equation}
    \mathbf{sty} = \sigma(\beta(t)) \oplus \sigma(\frac{\partial \beta(t) }{\partial t }) \oplus \sigma(\frac{\partial p(t) }{\partial t }),
\label{equ:sty}
\end{equation}
where $\oplus$ signifies the vector concatenation. %

To summarize, we make the following two conclusions: (1) one identity have multiple talking styles, (2) the talking style is closely related to the variance of facial and head pose movements inside each video, following which we define the style codes in Equation~\ref{equ:sty}. %
The style codes are utilized to synthesize diversified talking styles, details will be illustrated in the section~\ref{sec:method}. %





\section{Methodology}
\label{sec:method}

Following the style codes defined in section~\ref{sec:obv}, we propose a two-stage talking face synthesis framework to imitate arbitrary talking styles, as shown in Figure~\ref{fig:framework}. %
Our framework synthesizes stylized talking videos with the following three inputs: one static portrait of the speaker, the driven audio and the style reference video. 
In the first stage of our framework, we devise a latent-style-fusion~(LSF) model to synthesize stylized 3D talking faces through imitating arbitrary talking styles. %
Based on the synthesized 3D talking faces, in the second stage, we leverage the deferred neural render~\cite{thies2019deferred} and few-shot neural texture generation model to generate video frames photo-realistically. %
In the next two subsections, we will introduce the two stages respectively. %



\subsection{Stylized 3D Talking Face Synthesis}

In the first stage of our framework, we propose the latent-style-fusion~(LSF) model for talking face synthesis. %
Overall, the input of the LSF model is the driven audio and the reference talking video for style imitation. %
The LSF model learns motion related information from audio, and then combines latent audio representation with style information to synthesize 3D talking meshes with target talking style. %
Details will be illustrated as follows. %

For the driven input audio $\mathbf{X}_{a}$ of $T$ seconds, we firstly leverage the DeepSpeech~\cite{hannun2014deep} model to extract speech features. %
Deepspeech is a deep neural model for automatic speech recognition~(ASR). %
The extracted features from DeepSpeech not only contains rich speech information but also is robust to background noise and generalizes well to different identities. %
Feeding the input audio $\mathbf{X}_{a}$ to the DeepSpeech model, yields latent representations $\mathbf{X}_{d} \in \mathbb{R}^{50T \times D_{a}}$, %
where $D_{a}$ is the dimension of the DeepSpeech features, and $50T$ denotes that the DeepSpeech features have 50 frames per second. %
Afterwards, for the reference talking video, we calculate its style codes $\mathbf{sty} \in \mathbb{R}^{D_{s}}$ as illustrated in Section~\ref{sec:obv}  for style imitation, where $D_{s}$ is the dimension of the style codes. %

Having obtained the $\mathbf{X}_{d}$ and $\mathbf{sty}$, we now elaborate the 3D talking face synthesis process. %
We devise a latent-style-fusion~(LSF) model, which takes $\mathbf{X}_{d} \in \mathbb{R}^{50T \times D_{a}}$ and $\mathbf{sty} \in \mathbb{R}^{D_{s}}$ as input, outputs facial movements $\beta_{pred}(t) \in \mathbb{R}^{25T \times 64}$ and head pose movements $p_{pred}(t) \in \mathbb{R}^{25T \times 7}$ with 25 frames per second. %
Based on $\beta_{pred}(t)$ and $p_{pred}(t)$, we reconstruct the 3D talking meshes with the 3DMM face model~\cite{blanz1999morphable}. %

The LSF model leverages a latent fusion mechanism to both synthesize stylized faces and guarantee the synchronization between audio and motion. %
Specifically, as shown in Figure~\ref{fig:framework}, 
the LSF model firstly takes audio $\mathbf{X}_{d}$ as input and encodes $\mathbf{X}_{d}$ with the bottom part of ResNet-50~\cite{he2016deep}, yielding latent audio representation $\mathbf{X}_{l}$. %
Afterwards, the LSF model fuses the latent audio representation $\mathbf{X}_{l}$ and style codes $\mathbf{sty}$ to acquire mixed representation for synthesis. %
During the fusion process, the LSF model first dropouts the latent audio representation $\mathbf{X}_{l}$ to obtain $\mathbf{X}_{l}^{'}$, while the information from $\mathbf{sty}$ remains unchanged. %
Next, the LSF model concats each frame of $\mathbf{X}_{l}^{'}$ with style codes $\mathbf{sty}$, leading to the mixed representation. %
Further,  the LSF model leverages the top part of ResNet-50 to predict facial movements $\beta_{pred}(t)$ and head pose movements $p_{pred}(t)$ from the mixed representation. %
It's noteworthy to stress that the fusion between latent audio representation and style codes enables the synthesis of more expressive talking styles. Meanwhile, the dropout of audio information prevents from discarding style information for synthesis. %
The overall implementation of the LSF model is simple but effective. %

For the training stage of the LSF model, we adopt the parameter series $\beta(t), p(t)$ reconstructed from the 3D face reconstruction algorithm~\cite{deng2019accurate} as ground truth. %
For each training video, we calculate its style codes $\mathbf{sty}$ and randomly clip the the input audio $\mathbf{X}_{d}$ and the ground truth $\beta(t), p(t)$ to fixed length. %
Afterwards, we feed $\mathbf{X}_{d}$ and corresponding $\mathbf{sty}$ to the LSF model, yielding the predicted expression parameter series $\beta_{pred}(t)$ and $p_{pred}(t)$. %
Based on the predicted $\beta_{pred}(t)$ and $p_{pred}(t)$, we apply the $\mathrm{L_{1}}$ loss as follows:
\begin{equation}
    \mathcal{L}_{\mathrm{L_{1}}} = ||\beta(t) - \beta_{pred}(t)||_{1} + ||p(t) - p_{pred}(t)||_{1}.
\end{equation}
It is worth emphasizing that the training of the LSF model doesn't require any additional annotation on the speaking identity. %
Only by training on wild videos with stable talking styles, we obtain expressive style space. %


During the inference stage, feeding style codes of arbitrary talking videos to LSF model not only yields desired talking styles but also maintains the synchronization between the driven audio and talking faces. %
Meanwhile, we can interpolate among different talking styles to acquire new talking styles. %
For the audio representation $\mathbf{X}_{d}$ with arbitrary duration, since that the trained LSF model only digests the audio with fixed length, %
we apply the slide window strategy to synthesize the corresponding facial movements $\beta_{pred}(t)$ and head pose movements $p_{pred}(t)$. %

So far, we have obtained the untextured talking 3D faces. %
In the next subsection, we'll introduce how we render these 3D faces photo-realistically. %



\subsection{Photorealistic Render}



Conventional deferred neural render~\cite{thies2019deferred} requires substantial training data for each identity. In order to both synthesize photorealistic results and guarantee the few-shot capacity, we devise a few-shot neural texture generation model and combine the generated neural texture with the deferred neural render, which enables synthesizing photorealistic videos with only one source portrait. %
As shown in Figure~\ref{fig:framework}, the deferred neural render incorporates the generated neural texture, conducts UV texture sampling on the neural texture, and translates the sampled image to photorealistic domain. %

\begin{algorithm}
  \caption{UV Texture Sampling}
  \label{alg:sample}
  \begin{algorithmic}[1]
    \State $\mathbf{X}_{uv} \in \mathbb{R}^{2\times H \times W}$
    \State $\mathbf{Y}_{t} \in \mathbb{R}^{D_{t}\times H_{t} \times W_{t}}$
    \State $\mathbf{X}_{s} \in \mathbb{R}^{D_{t}\times H \times W}$
    \For {$i \gets 1$ to $H$}
        \For {$j \gets 1$ to $W$}    
            \State $u = \mathbf{X}_{uv}[0, i, j]$
            \State $v = \mathbf{X}_{uv}[1, i, j]$
            \State $\mathbf{X}_{s}[:, i, j]$ = BilinearSample$(\mathbf{Y}_{t}, u, v)$
        \EndFor
    \EndFor
\State \Return $\mathbf{X}_{s}$
  \end{algorithmic}
\end{algorithm}

Detailedly, for the input 3D talking faces, we firstly leverage the UVAtlas tool~\footnote{https://github.com/microsoft/UVAtlas} to obtain the UV coordinate of each vertex in the 3D model. %
Afterwards, we rasterize the 3D face model to 2D image $\mathbf{X}_{uv} \in \mathbb{R}^{2 \times H \times W}$, of which each pixel represents the UV coordinate. %
Subsequently, for the input portrait $\mathbf{X}_{p} \in \mathbb{R}^{3\times H\times W}$ and the 3D face model, we extract the RGB texture $\mathbf{X}_{t} \in \mathbb{R}^{3 \times H_{t} \times W_{t}}$, where $H_{t}, W_{t}$ signifies the height and width of the texture. %
Based on $\mathbf{X}_{t}$, we leverage the pix2pix~\cite{pix2pix2017} model to generate neural texture $\mathbf{Y}_{t} \in \mathbb{R}^{D_{t} \times H_{t} \times W_{t}}$, where $D_{t}$ signifies the dimension of the neural texture. With the neural texture, we conduct UV texture sampling on $\mathbf{X}_{uv}$ to obtain the sampled image $\mathbf{X}_{s} \in \mathbb{R}^{D_{t} \times H \times W}$, the details of the sampling algorithm are illustrated in the Algorithm~\ref{alg:sample}. %
Finally, we translate the sampled image $\mathbf{X}_{s}$ to photorealistic image through the pix2pixHD~\cite{wang2018high} model. %

During the training stage, the few-shot texture generation model and the deferred neural render model are trained simultaneously. %
Given the rasterized input $\mathbf{X}_{uv}$, we denote the rendered image as $\mathbf{Y}^{'}$ and the ground truth image as $\mathbf{Y}$. %
We combine the perceptual loss~\cite{johnson2016perceptual} and $\mathrm{L_{1}}$ loss together as $\mathcal{L}$ to optimize the neural texture and pix2pixHD model. Formally: %

\begin{equation}
    \mathcal{L} = ||\mathbf{Y} - \mathbf{Y}^{'}||_{1} + ||\phi(\mathbf{Y}) - \phi(\mathbf{Y}^{'})||_{1},
\end{equation}
where $\phi(\cdot)$ is the first few layers of VGGNet~\cite{DBLP:journals/corr/SimonyanZ14a} pretrained on the ImageNet~\cite{deng2009imagenet}. %
Due to the limitation of the 3DMM face model, in our render, %
we only synthesize the facial part of the image, without considering the hair and background rendering. %

\section{Experiments}
In this section, we conduct extensive experiments to demonstrate the effectiveness of our framework. %
We evaluate our framework on the collected \textit{Ted-HD} dataset. %
Our method has acquired better synthesis results both qualitatively and quantitatively. %

\subsection{Dataset}
As illustrated in Section~\ref{sec:obv}, the currently available datasets are either collected in lab which have constrained talking styles or collected in the wild of which the talking styles are unstable and noisy. %
Therefore for the training and testing of the LSF model, we leverage the \textit{Ted-HD} dataset described in Section~\ref{sec:obv}. %
Totally, the \textit{Ted-HD} dataset has 834 clips of videos, we select 799 clips for training, and hold out the remained 35 for testing. %
The training set and test set have no overlap on identities. %
Additionally, for the training of the deferred neural render and the  few-shot neural texture generation model, we leverage the Lip Reading in the Wild~(LRW)~\cite{Chung16} dataset. %

\subsection{Implementation Details}

During the training of the LSF model, the input DeepSpeech audio features have 50 frames per second~(FPS), while each frame has the dimension $D_{a}$ of 29. The input style codes $\mathbf{sty}$ have the dimension $D_{s}$ of 135~(64 for $\sigma(\beta(t))$, 64 for $\sigma(\frac{\partial \beta(t) }{\partial t })$, 7 for $\sigma(\frac{\partial p(t) }{\partial t })$). %
The predicted facial movements $\beta_{pred}(t)$ and the head pose movements $p_{pred}(t)$ have 25 frames per second. %
For the convenience of training, we randomly clip the input to 80 frames, and clip the output to 32 frames. %
For the implementation of the LSF model, we apply the ResNet-50 as the backbone. We leverage the first 16 layers of ResNet-50 to encode the DeepSpeech features, combine the encoded features with style codes, and leverage the last 34 layers of ResNet-50 to predict the motion series. %
When optimizing, we adopt the Adam optimizer~\cite{DBLP:journals/corr/KingmaB14} to train the LSF model with the initial learning rate of $5\times 10^{-4}$. %
We train for 50000 iterations with a mini-batch size of 128 samples. %

During the training of the deferred neural render and the few-shot neural texture generation model, the input UV image $\mathbf{X}_{uv}$ has a size of $2\times224\times224$, while the neural texture has a size of $16\times64\times64$. %
The dimension $D_{t}$ of the texture is set to 16, which enables each pixel to contain richer texture information. Meanwhile, the texture size is set to be smaller than the UV image size, which avoids oversampling in the sample process. The output image $\mathbf{X}_{s}$ is conventional RGB image with size of $3\times224\times224$. %
When optimizing, we adopt the Adam optimizer to simultaneously train the neural render and the texture generation model. %
The learning rate is set to $2\times 10^{-4}$. %
We train for 1000000 iterations with mini-batch size of 6 samples. %

\begin{figure*}[h]
  \centering
  \includegraphics[width=0.8\linewidth]{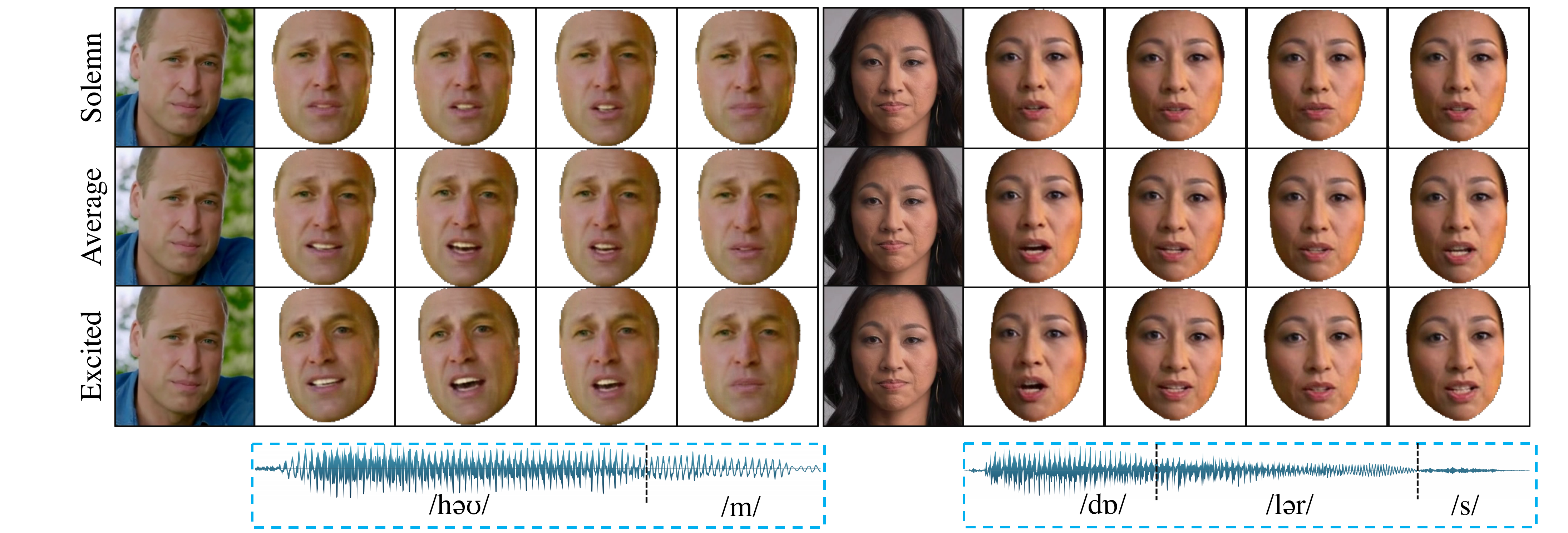}
  \caption{The interpolation results between excited talking style and solemn talking style with the same driven audio. The first row has solemn talking style, and the last row has excited talking style, while the middle row has the average style between the first and the last. Among these rows, we observe smooth transition of facial and head pose movements. }
  \label{fig:interpolate}
\end{figure*}

\subsection{Comparison with VOCA on Style Synthesis}


To the best of our knowledge, the VOCA model~\cite{cudeiro2019capture} is the only available method that captures diversified talking styles of facial movements. %
Therefore, in this section, we systematically compare our method with the VOCA model~\cite{cudeiro2019capture} to demonstrate the effectiveness of the LSF model. %
Different from our method, the VOCA model learns identity-level talking styles. %
Specifically, the VOCA model injects a one-hot identity code to the time convolution network, and directly predicts face model vertexes from the DeepSpeech features and the identity code. %
By adjusting the one-hot identity code, the VOCA model outputs different talking styles. %

\begin{table}[h]
\caption{The mean opinion scores~(MOS) of different metrics, higher signifies better. TSE denotes talking style expressiveness, FMN denotes facial movement naturalness, and AVS denotes audio-visual synchronization. }
\begin{tabular}{c|c c c}
\hline
     & TSE  & FMN  & AVS  \\ \hline
VOCA~\cite{cudeiro2019capture} & 3.28 & 3.13 & 3.56 \\
LSF  & 3.41 & 3.21 & 3.64 \\ \hline
\end{tabular}
\label{tab:com_voca_lsf}
\end{table}

We compare the style space learned from the VOCA model and the style space learned from the LSF model with extensive user studies. %
Specifically, we randomly select 10 clips of driven audio, each with a duration from 10 to 20 seconds. %
Afterwards, we randomly sample 5 talking styles from the VOCA style space and 5 talking styles from the style space of our method. %
With the sampled talking styles and the driven audio, we synthesize corresponding talking faces and retarget the synthesized faces to the same identity. %
Afterwards, we subsume videos with the same driven audio and synthesis model to the same group. %
For each group of video, we invite 20 participants to rate (1) the expressiveness of the talking style, (2) the naturalness of facial movements, (3) the audio-visual synchronization between driven audio and talking face. %
We ask participants to rate the mean opinion score~(MOS)~\cite{recommendation2006vocabulary} in the range 1-5~(higher MOS score denotes better results). %
When showing videos to participants, since that the VOCA model only gives untextured 3D faces, we also just give the 3D talking faces synthesized from the LSF model, without utilizing the deferred neural render to generate photorealistic results for fair comparison. %

Table~\ref{tab:com_voca_lsf} demonstrates the results of user studies. %
Through the experimental results, we observe that our LSF model has acquired higher talking style expressiveness, which verifies that the style space learned by taking style imitation is more expressive than the identity-level style space. %
Meanwhile, our LSF model outperforms the VOCA model by 0.08 in terms of the facial movement naturalness and audio-visual synchronization. %
Such results confirm the effectiveness of style imitation in LSF model further. %
Additionally, compared with the VOCA model which requires substantial training data for each identity, our method is trained on wild dataset which does not require any annotation on identity or talking style. %


\subsection{Study on the Style Space}
In this section, we extensively investigate the style space learned in our method. %
We conduct two qualitative experiments, which verify that not only the synthesized talking styles are diversified, but new talking styles can also be generated from the interpolation of different talking styles. 

\begin{figure}[h]
  \centering
  \includegraphics[width=0.7\linewidth]{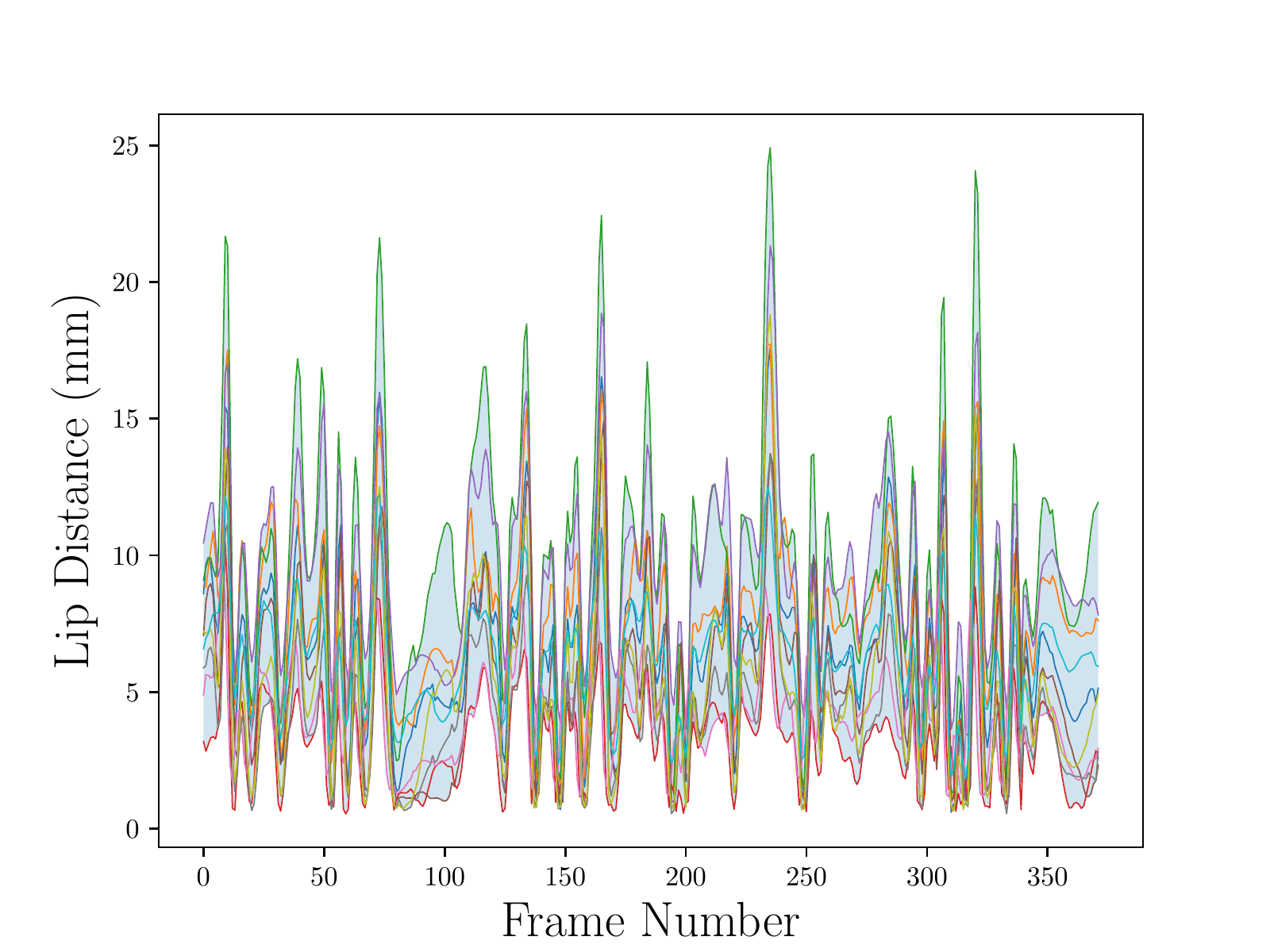}
  \caption{The distance between the lower lip and the upper lip for the same driven audio conditioned on different talking styles. Different color denotes different talking styles. }
  \label{fig:lipdis}
  \vspace{-5pt}
\end{figure}

To verify the diversification of the talking styles, we visualize the distance between lower and upper lip as a function of time. %
Specifically, we randomly select one piece of driven audio and 10 different talking styles, %
and then synthesize 10 facial movements corresponding to each talking style and driven audio. %
Afterwards, we calculate the lip distance as shown in Figure~\ref{fig:lipdis}. %
Through Figure~\ref{fig:lipdis} we observe that the lip distance significantly varies among different talking styles, which verifies that the LSF model is able to synthesize diversified talking styles. %
Meanwhile, different talking styles demonstrate similar trends of fluctuation, the peak and the valley of the distance curve highly overlap, which confirms that the LSF model also guarantees the synchronization between audio and synthesized motions. %

\begin{figure*}[h]
  \centering
  \includegraphics[width=0.87\linewidth]{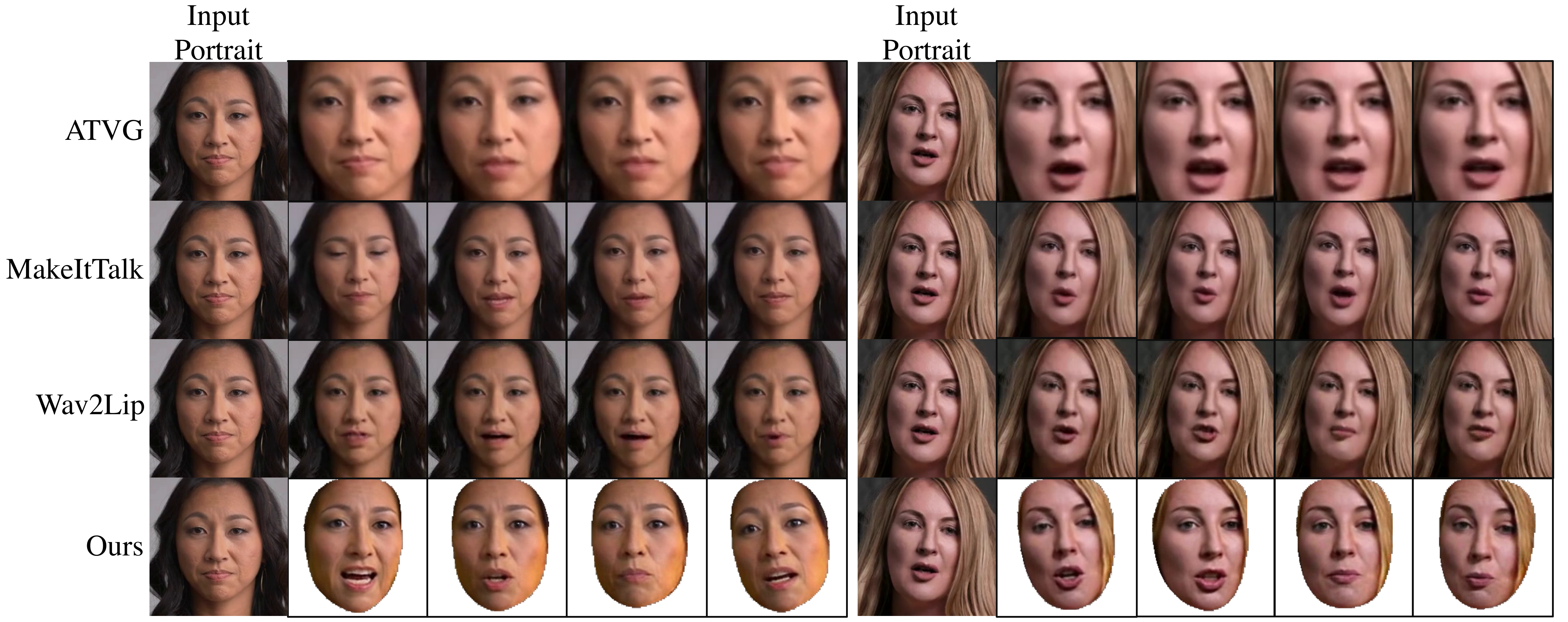}
  \caption{Comparison with several baseline methods~(ATVG, MakeItTalk and Wav2Lip). Our method not only yields expressive facial and head pose movements, but also synthesizes photorealistic videos. }
  \label{fig:one_compare}
\end{figure*}

To confirm that the style space in the LSF model is expressive and interpolative, we visualize the interpolation results of different talking styles, as shown in Figure~\ref{fig:interpolate}. %
Detailedly, we select two representative talking styles: excited and solemn, and conduct linear interpolation between the two styles to generate new talking styles. %
From each row of Figure~\ref{fig:interpolate}, we observe that the facial and head pose movements translate smoothly from excited talking style to solemn style. %
For the excited talking style, the lip motion is exaggerated and the head frequently shakes. %
Meanwhile, for the solemn talking style, the lip motion and head pose are stable. %
We provide more synthesis results in the supplementary materials. %


\subsection{Comparison with One-Shot Synthesis}
\label{sec:cmp}

In this section, we conduct experiments to demonstrate that our method synthesizes more natural and expressive talking faces compared with several baseline methods. %
Specifically, we compare our method with the following baseline methods: (1) the ATVG framework~\cite{chen2019hierarchical}, (2) the MakeItTalk framework~\cite{zhou2020makelttalk}, (3) the Wav2Lip framework~\cite{prajwal2020lip}. %
For Wav2Lip framework which requires few seconds of videos as input, %
we repeat input portrait as videos for fair comparison. %
Figure~\ref{fig:one_compare} shows some synthesis results %
We observe that our method has more expressive facial movements and head pose movements while also guarantees the synthesized results to be photorealistic. 

Additionally, we conduct both user studies and quantitative evaluations on the \textit{Ted-HD} dataset to verify the effectiveness of our method. %
Specifically, for the user studies, %
we firstly synthesize videos with the randomly selected 20 clips of driven audios and 5 different identities. %
Afterwards, for each video, %
we invite 20 participants to rate (1) the naturalness of facial movements, (2) the audio-visual synchronization between driven audio and talking face. %
The mean opinion score~(MOS) is rated in the range 1-5. %
Besides, we also evaluate the synthesized video quality with the signal-noise-ratio~(SNR) metric. %
We do not leverage the PSNR metric since that the ground truth talking videos with arbitrary talking styles are not available. %

\begin{table}[h]
\caption{Comparison with baseline methods on the \textit{Ted-HD} dataset, where FMN denotes facial movement naturalness, and AVS denotes audio-visual synchronization. The Pre-Fusion method removes latent style fusion in the LSF model, details are illustrated in Section~\ref{sec:ablation}}
\begin{tabular}{c|c c c}
\hline
           & FMN  & AVS  & SNR~(dB)  \\ \hline
ATVG~\cite{chen2019hierarchical}       & 2.71 & 3.51 & 2.98 \\
MakeItTalk~\cite{zhou2020makelttalk} & 3.08 & 3.13 & 3.01 \\
Wav2Lip~\cite{prajwal2020lip}    & 2.97 & 4.28 & 2.78 \\ \hline
Pre-Fusion  & 2.19 & 2.25 & 5.70 \\
Ours       & 3.59 & 3.75 & 5.76 \\ \hline
\end{tabular}
\label{tab:com_baseline}
\end{table}

\vspace{-0.75mm}

Table~\ref{tab:com_baseline} shows the comparison results. %
From Table~\ref{tab:com_baseline} we observe that our method has achieved the most expressive facial motion and best video quality. %
We also notice that the Wav2Lip method achieves unsatisfying motion naturalness since that it cannot resolve the one-shot synthesis scenario. %
Meanwhile, we observe that the AVS of our method is slightly lower than Wav2Lip, that is because the talking style synthesis in our LSF model mildly sacrifices the performance of audio-visual synchronization, but our method still has better AVS performance compared with ATVG and MakeItTalk. %


\subsection{Effectiveness of Latent Style Fusion}
\label{sec:ablation}
To verify the rationality of the latent style fusion mechanism in LSF model, we conduct the following ablation experiments. %
For comparison, we remove the latent style fusion mechanism, and directly concatenate DeepSpeech representation with style codes as the input of ResNet-50. %
Afterwards, we compare the synthesized results with LSF model through similar user studies as Section~\ref{sec:cmp} does. %
The experimental results are shown in the last two rows of Table~\ref{tab:com_baseline}. %
From the results we observe that the motion naturalness and audio-visual synchronization would degrade significantly once we remove the latent style fusion mechanism, which verifies the effectiveness of the latent style fusion. %
Meanwhile, the video quality remains constant since that the motion synthesis does not influence the photorealistic render stage. %









\section{Conclusion}

In this paper, we propose the concept of style imitation for audio-driven talking face synthesis. %
To imitate arbitrary talking styles, we firstly formulate the style codes of each talking video as several interpretable statistics of 3DMM parameters. %
Afterwards, we devise a latent-style-fusion~(LSF) model to synthesize stylized talking faces according to the style codes and driven audio. %
The incorporation of style imitation not only circumvents the annotation for talking style during the training phase, but also endows the capacity of arbitrary style synthesis and new talking style generation. %
Additionally, to synthesize expressive talking styles, we collect \textit{Ted-HD} dataset with 834 clips of talking videos, which contains stable and diversified talking styles. %
We conduct extensive experiments on the constructed dataset and obtain expressive synthesis results with our \textit{Ted-HD} dataset and LSF model. %
The constructed \textit{Ted-HD} dataset will be made publicly available in the future. %
We hope that the proposal of talking style imitation and the construction of \textit{Ted-HD} dataset pave a new way for audio-driven talking face synthesis. %

\section{Acknowledgments}

This work is supported by the National Key R\&D Program of China under Grant No. 2020AAA0108600, the state key program of the National Natural Science Foundation of China (NSFC) (No.61831022), Beijing Academy of Artificial Intelligence No. BAAI2019QN0302.

\newpage

\bibliographystyle{ACM-Reference-Format}
\balance
\bibliography{sample-base}










\end{document}